# MAP Complexity Results and Approximation Methods

James D. Park
Computer Science Department
University of California
Los Angeles, CA 90095
jd@cs.ucla.edu

## Abstract

MAP is the problem of finding a most probable instantiation of a set of variables in a Bayesian network, given some evidence. MAP appears to be a significantly harder problem than the related problems of computing the probability of evidence (Pr), or MPE (a special case of MAP). Because of the complexity of MAP, and the lack of viable algorithms to approximate it, MAP computations are generally avoided by practitioners.

This paper investigates the complexity of MAP. We show that MAP is complete for $NP^{PP}$. We also provide negative complexity results for elimination based algorithms. It turns out that MAP remains hard even when MPE, and Pr are easy. We show that MAP is NP-complete when the networks are restricted to polytrees, and even then can not be effectively approximated.

Because there is no approximation algorithm with guaranteed results, we investigate best effort approximations. We introduce a generic MAP approximation framework. As one instantiation of it, we implement local search coupled with belief propagation (BP) to approximate MAP. We show how to extract approximate evidence retraction information from belief propagation which allows us to perform efficient local search. This allows MAP approximation even on networks that are too complex to even exactly solve the easier problems of computing Pr or MPE. Experimental results indicate that using BP and local search provides accurate MAP estimates in many cases.

## 1 Introduction

The task of computing the Maximum a Posteriori hypothesis (MAP) is to find the most likely configuration of a set of variables (which we will call the MAP variables) given (partial) evidence about the complement of that set (the non-MAP variables).

One specialization of MAP which has received a lot of attention is the Most Probable Explanation (MPE). MPE is the problem of finding the most likely configuration of a set of variables given an evidence instantiation for the complement of that set. The primary reason for this attention is that MPE seems to be a much simpler problem than its MAP generalization.

Unfortunately, MPE is not always suitable for the task of providing explanations. For example, in system diagnosis, where the health of each component is represented as a variable, one is interested in finding the most likely configuration of health variables only – the likely input and output values of each component are not of interest. Additionally, the projection of an MPE solution on these health variables is usually not a most likely configuration. Nor is the configuration obtained by choosing the most likely state of each variable separately.

MAP turns out to be a very difficult problem even when compared to MPE or computing the probability of evidence (Pr). In section 2 we present some complexity results for MAP that indicate that neither exact nor approximate solutions can be guaranteed, even under very restricted circumstances. Still, MAP remains an important problem, and one we would like to be able to generate solutions for. Our approach is to provide best effort approximation methods. In section 3 we discuss a general approach to approximating MAP and provide one instantiation of that approach that is based on belief propagation and local search, which allows MAP approximations even when exact MPE and Pr computations are not feasible.

## 2 MAP Complexity

In this section, we begin by reviewing some complexity theory classes and terminology that pertain to the complexity of MAP. Next, we examine the complexity of MAP in the general case. We then examine the complexity of current state of the art MAP algorithms based on variable elimination. We conclude the complexity section by examining the complexity of MAP on polytrees.

### 2.1 Complexity Review

We assume that the reader is familiar with the basic notions of complexity theory like the hardness and completeness of languages, as well as the complexity class NP. For an in-depth introduction to complexity theory see [13].

In addition to NP, we will also be interested in the class PP and a derivative of it. Informally, PP is the class which contains the languages for which there exists a nondeterministic Turing machine where the majority of the nondeteriministic computations accept if and only if the string is in the language. PP can be thought of as the decision version of the functional class #P. As such, PP is a powerful language. In fact NP $\subseteq$ PP, and the inequality is strict unless the polynomial hierarchy collapses to the second level.[1]

Another idea we will need is the concept of an oracle. Sometimes it is useful to ask questions about what could be done if an operation were free. In complexity theory this is modeled as a Turing machine with an oracle. An oracle Turing machine is a Turing machine with the additional capability of being able to obtain answers to certain queries in a single time step. For example, we may want to designate the class of languages that could be recognized in nondeterminstic polynomial time if any PP query could be answered for free. The class of languages would be NP with a PP oracle, which is denoted NP$^{PP}$.

In this paper, we will be dealing with the decision versions of the problems. For example, the decision problem for MAP is: *Given a Bayesian Network with rational parameters, a subset of its variables X, evidence e (which consists of a partial instantiation of the non-MAP variables) and a rational threshold k, is there an instantiation x of X such that $Pr(x,e) > k$?* The decision problems for MPE and Pr are defined similarly.

### 2.2 MAP Complexity for the General Case

Computing MPE, Pr, and MAP are all NP-Hard, but there still appears to be significant differences in their complexity. MPE is basically a combinatorial optimization problem. Computing the probability of a complete instantiation is trivial, so the only real difficulty is determining which instantiation to choose. MPE is NP-complete.[2] Pr is a completely different type of problem, characterized by counting instead of optimization and is PP-complete [7] (notice that this is the complexity of the decision version, not the functional version which is #P-complete [17]). MAP combines both the counting and optimization paradigms. In order to compute the probability of a particular instantiation, an inference query is needed. Optimization is also required, in order to be able to decide between the many possible instantiations. This is reflected in the complexity of MAP.

**Theorem 1** *MAP is* NP$^{PP}$-*complete.*[3]

*Proof:* Membership in NP$^{PP}$ is immediate. Given any instantiation x of the MAP variables, we can verify if it is a solution by querying the PP oracle if $Pr(x,e) > k$.

To show hardness, we reduce E-MAJSAT [10] (the canonical SAT oriented complete problem for NP$^{PP}$) to MAP. E-MAJSAT is defined as follows: *Given a Boolean formula $\phi$ over n variables $x_1, ..., x_n$, and an integer k, $1 \leq k \leq n$, is there an assignment to the first k variables such that the majority of the assignments to the remaining $n - k$ variables satisfy $\phi$?* First, we create a Bayesian Network that models the Boolean expression. For each variable in the expression, we create an analogous variable in the network with uniform prior probability. Then, for each logical operator, we create a variable whose parents are the variables corresponding to its operands, and whose CPT encodes the truth table for that operator (see Figure 1 for a simple example). Let $v_\phi$ be the network variable corresponding to the top level operand. For a particular instantiation x of variables $x_1, ..., x_k$, we let e = $\{v_\phi = T\}$. Then,

$$Pr(\mathbf{x}, \mathbf{e}) = \sum_{x_{k+1},...,x_n} Pr(\mathbf{x}, x_{k+1}, ..x_n, v_\phi = T)$$

---

[1]This is a direct result of Toda's theorem [20]. From Toda's theorem P$^{PP}$ contains the entire polynomial hierarchy (PH), so if NP = PP, then PH $\subseteq$ P$^{PP}$ = P$^{NP}$.

[2]The NP-hardness of the functional version of MPE was shown in [19]. We are not aware of any published proof of the completeness of the decision problem, so we sketch it here. Membership is immediate, since the score for a purported solution can be tested in linear time. Hardness is based on using the standard Bayesian network simulation of a Boolean expression (c.f. Theorem 1) to solve SAT. MPE($v_\phi = T$) > 0 if and only if the expression is satisfiable.

[3]This result was stated without proof in [9]. The author attributed the result [8] to an unpublished proof by Mark Peot.

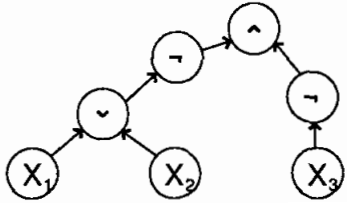

Figure 1: The Bayesian network produced using the reduction in Theorem 1 for Boolean formula $\neg(x_1 \vee x_2) \wedge \neg x_3$.

$$= (\#\text{satisfied})/2^n$$

Since there are $2^{n-k}$ possible instantiations of $x_{k+1}, ..., x_n$, we have

$$Pr(\mathbf{x}, \mathbf{e}) = (\text{fraction satisfied})/2^k$$

So the MAP query over variables $x_1, ..., x_k$ with evidence $v_\phi = T$ and threshold $1/2^{k+1}$ is true if and only if the E-MAJSAT query is also true. □

This class has also been shown to be important in probabilistic planning problems [10].

NP$^{\text{PP}}$ is a powerful class, even compared to NP and PP. They are related by NP $\subseteq$ PP $\subseteq$ NP$^{\text{PP}}$, where the equalities are considered very unlikely. In fact, NP$^{\text{PP}}$ contains the entire polynomial hierarchy [20].

Additionally, because MAP generalized Pr, MAP inherits the wild nonapproximability of Pr shown in [17].

**Corollary 2** *For any $\epsilon > 0$, approximating MAP within $2^{n^{1-\epsilon}}$ is NP-hard where $n$ is the number of variables in the network.*

So, if P $\neq$ NP then no polynomial time algorithm exists for approximating MAP that can guarantee subexponential relative error.

### 2.3 Results for Elimination Algorithms

Solution to the general MAP problem seems out of reach, but what about for "easier" networks? State of the art exact inference algorithms (variable elimination [4], join trees [6, 18, 5], recursive conditioning [2]) can compute $Pr(\mathbf{e})$ and MPE in space and time complexity that is exponential only in the width of the elimination order used. This allows many networks to be solved using reasonable resources even though the general problems are very difficult. Similarly, state of the art MAP algorithms can solve MAP with time and space complexity that is exponential only in width of the elimination order. Unfortunately, for MAP, not all orders can be used. In practice the order is generally generated by restricting the elimination order to eliminate all of the MAP variables last. This tends to produce elimination orders with widths much larger than those available for Pr and MPE, often placing exact MAP solutions out of reach [15]. We now consider the question of whether there are less stringent conditions for valid elimination orders, that may allow for orders with smaller widths.

Elimination algorithms exploit the fact that summation commutes with summation, and maximization commutes with maximization in order to essentially factor the problem. Given an ordering, elimination algorithms work by stepping through the ordering, collecting the potentials mentioning the current variable, multiplying them, then replacing them with the potential formed by summing out (or maximizing) the current variable from the product. This process can be thought to induce an *evaluation tree*. The evaluation tree for an order consists of the potentials generated by performing the variable elimination, where an edge means that the child was one of the potentials that were combined to form the parent (see Figure 2 for an example). The width of the elimination order is the size (measured in the number of variables) of the largest potential in the evaluation tree.

Because maximization and summation do not commute, not all variable orders generate evaluations that are valid. That is, trying to perform elimination using some orders will produce incorrect results. MAP requires that summation be performed before maximization. Thus, the criteria that needs to be satisfied is that a potential cannot be maximized if it mentions any summation variables. An elimination order is *valid* (because it generates a valid evaluation tree) if the induced evaluation tree never maximizes a variable out of a potential that mentions a summation variable. The standard way of ensuring a valid order is to eliminate all of the summation variables before any of the maximization variables. Two questions present themselves. First, are there valid orderings that interleave summation and maximization variables? And second, if so, can they produce widths smaller than those generated by eliminating all summation variables, then all maximization variables?

The answer to the first question is yes, there are other valid elimination orders. To see that, we introduce the notion of the elimination tree. An elimination order induces an *elimination tree* which consists of the variables of the order, where an edge from parent to child indicates that the potential generated by eliminating the child was combined to form the potential generated by eliminating the parent. The elimination tree can be thought of as a high level summary of the evaluation tree. Figure 2 shows a sample network and elimination order, with its associated evaluation tree and elimination tree. The elimination tree defines a partial ordering of the variables where a child in the tree must be eliminated before the parent. Any elimination order that obeys the partial order induces the same evaluation tree. Thus, if an order is valid, all other orders that share the same elimination tree are also valid. Additionally, since they share the same evaluation tree, they all have the same width. Figure 2 shows the tree induced by using the order AEBDC (which eliminates summation variables first) to solve MAP(C,D). An equivalent order that interleaves summation and maximization variables is ABDEC. Typically, there are many valid interleaved elimination orders. Unfortunately, allowing interleaved orders does not help.

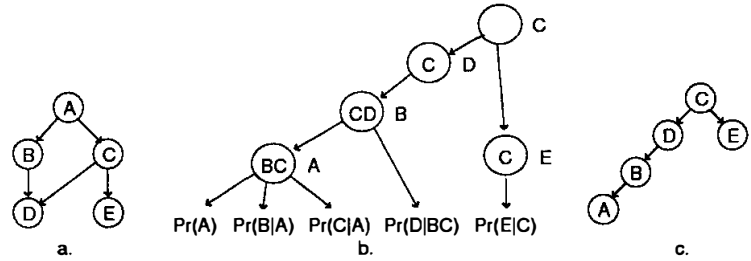

Figure 2: A Bayesian network (a), an evaluation tree (b), and an elimination tree (c) corresponding to the elimination order ABEDC.

**Theorem 3** *For any valid MAP elimination order, there is an ordering of the same width in which all of the maximization variables are eliminated last.*

*Proof:* Consider the elimination tree induced by any valid elimination order. No summation variable is the parent of any maximization variable. This can be seen by considering any maximization variable. When the corresponding potential was maximized, it had no summation variables, and so the resulting potential also has no summation variables. Hence, any parent of a maximization variable must also be a maximization variable. Since no summation variable is a parent of a maximization variable, all summation variables can be eliminated first in any order consistent with the partial order defined by the elimination tree. Then, all the maximization variables can be eliminated, again obeying the partial ordering defined by the elimination tree. Because the produced order has the same elimination tree as the original order, they have the same width. □

### 2.4 MAP on Polytrees

Theorem 3 has significant complexity implications for elimination algorithms even on polytrees.

**Theorem 4** *Elimination algorithms require exponential resources to perform MAP, even on some polytrees.*

*Proof:* Consider computing MAP($X_1...X_n, \{S_n = T\}$) for a network consisting of variables $X_1, ..., X_n, S_0, ..., S_n$ with topology as shown in Figure 3. By Theorem 3, there is no order better than eliminating all of the non-MAP variables. But, after the non-MAP variables are eliminated, all of the MAP variables appear in a single potential. Thus the width is linear in the number of variables, and the algorithm requires exponential resources. □

Which variables are maximized makes a crucial difference in the complexity of MAP computations. For example, the problem of maximizing over $X_1...X_{n/2}, S_0...S_{n/2}$ instead of $X_1...X_n$ can be solved in linear time.

It turns out that finding a good general algorithm for MAP on polytrees is unlikely.

**Theorem 5** *MAP is NP-Complete when restricted to polytrees.*

*Proof:* Membership is immediate. Given a purported solution instantiation **x**, we can compute Pr(**x**,**e**) in linear time and test it against the bound. To show hardness, we reduce MAXSAT to MAP on a polytree. A similar reduction was used in [10] and [14]. The MAXSAT problem is defined as follows: *Given a set of clauses $C_1, ..., C_m$ over variables $x_1, ..., x_n$ and an integer bound k, is there an assignment of the variables, such that more than k clauses are satisfied.* The idea behind the reduction is to model randomly selecting a clause, then successively checking whether the instantiation of each variable satisfies the selected clause. The clause selector variable $S_0$ with possible values $1, 2, ..., m$ has a uniform prior. Each propositional variable $x_i$ induces two network variables $X_i$ and $S_i$. $X_i$ represents the value of $x_i$, and has a uniform prior. $S_i$ represents whether any of $x_1, ..., x_i$ satisfy the selected clause. $S_i = 0$ indicates that the selected clause was satisfied by one of $x_1, ..., x_i$. $S_i = c > 0$

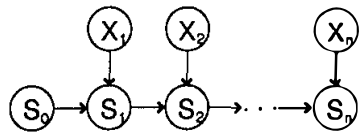

Figure 3: The network used in the reduction of Theorem 5.

indicates that the selected clause $C_c$ was not satisfied by $x_1, ..., x_i$. The parents of $S_i$ are $x_i$ and $S_{i-1}$ (the topology is shown in Figure 3). The CPT for $S_i$, for $i \geq 1$ is defined as

$$Pr(S_i|X_i, S_{i-1}) = \begin{cases} 1 & \text{if } S_i = S_{i-1} = 0 \\ 1 & \text{if } S_i = 0 \text{ and } S_{i-1} = j, \text{ and} \\ & X_i \text{ satisfies } C_j \\ 1 & \text{if } S_i = S_{i-1} = j \text{ and } X_i \text{ does} \\ & \text{not satisfy } C_j \\ 0 & \text{otherwise} \end{cases}$$

In words, if the selected clause was not satisfied by the first $i-1$ variables ($S_{i-1} \neq 0$), and the $x_i$ satisfies it, then $S_i$ becomes satisfied ($S_i = 0$) otherwise, $S_i = S_{i-1}$. Then for a particular instantiation c of $S_0$ and x of $X_1, ..., X_n$, $Pr(c, x, S_n = 0) = 1/(m2^n)$ if x satisfies clause $C_c$, 0 otherwise. Thus MAP over $X_1, ..., X_n$ with evidence $S_n = 0$ and bound $k/(m2^n)$ solves the MAX-SAT problem as well. □

Additionally, because MAX-SAT is MAXSNP-complete, we have the following corollary:

**Corollary 6** *MAP on polytrees is* MAXSNP-*hard*.

This means that there is no polynomial time approximation scheme for MAP on polytrees unless P = NP.

## 3 Approximating MAP

The complexity of MAP places exact solution out of reach in all but the simplest cases. Good approximations can not be guaranteed either. Still, we want some method to generate at least approximate solutions to the problem. Typically practitioners resort to computing individual posteriors, or computing MPE, and projecting the solution onto the MAP variables. Unfortunately both methods in general produce relatively poor approximations to MAP. Also, when the network is complex, both methods are too complicated to compute exactly.

We propose a general framework for approximating MAP. MAP consists of two problems that are hard in general — optimization and inference. A MAP approximation algorithm can be produced by substituting approximate versions of either the optimization or inference component (or both). The optimization problem is defined over the MAP variables, and the score for each solution candidate instantiation s of the MAP variables is the (possibly approximate) probability $Pr(s, e)$ produced by the inference method. This allows solutions tailored to the specific problem. For networks whose treewidth is manageable, but contains a hard optimization component (e.g. the polytree examples discussed previously), exact structural inference can be used, coupled with an approximate optimization algorithm. Alternatively, if the optimization problem is easy (e.g. there are few MAP variables) but the network isn't amenable to exact inference, an exact optimization method could be coupled with an approximate inference routine. If both components are hard, both the optimization and inference components need to be approximated.

The only previous algorithms for approximating MAP of which we are aware are instantiations of this framework. They both use an exact probability engine, but an approximate optimization engine [3, 15], and so are feasible for networks amenable to exact inference.

We now present a new MAP approximation algorithm which uses local search to approximate the optimization component, and belief propagation to approximate the inference component. This extends the realm of problems where MAP approximations can be effectively generated to problems that can be approximated well by belief propagation.

Belief propagation has a number of qualities that make it a good candidate to use as the approximate probability engine for MAP approximation. Experimental results have shown impressive performance in a variety of domains. It has effective methods for computing MPE and posteriors of individual nodes, which are both powerful initialization methods for the local search. Recent work [21] has demonstrated how to use BP to estimate $Pr(e)$, which is the primary requirement for using it as a subroutine to MAP. Also, approximate retracted marginals $Pr(x|e - X)$ can be computed locally for each variable. The notation $e - X$ represents the instantiation formed by removing the assignment of $X$ from e. The ability to approximate retracted marginals provides a linear speed up for the search.

### 3.1 Belief Propagation Review

Belief propagation was introduced as an exact inference method on polytrees [16]. It is a message passing algorithm in which each node in the network sends a message to its neighbors. These messages, along with the CPTs and the evidence can be used to compute posterior marginals for all of the variables. In net-works with loops, belief propagation is no longer guaranteed to be exact and successive iterations generally produce different results, so belief propagation is typically run until the message values converge. This has been shown to provide very good approximations for a variety of networks [11, 12], and has recently received a theoretical explanation [22].

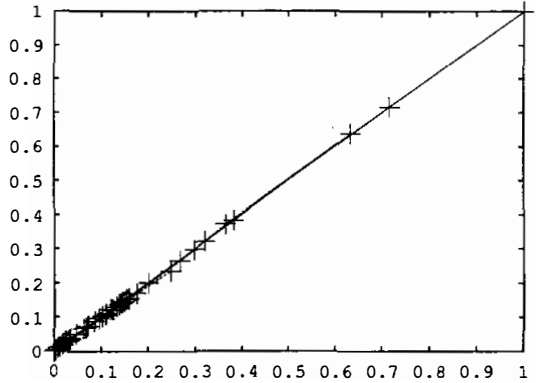

Figure 4: A scatter plot of the exact versus approximate retracted values of 30 variables of the Barley network.

Belief propagation works as follows. Each node $X$, has an evidence indicator $\lambda_X$ where evidence can be entered. If the evidence sets $X = x$, then $\lambda_X(x) = 1$, and is 0 otherwise. If no evidence is set for $X$, then $\lambda_X(x) = 1$ for all $x$. After evidence is entered, each node $X$ sends a message to each of its neighbors. The message a node $X$ with parents $\mathbf{U}$ sends to child $Y$ is computed as

$$M_{XY} = \alpha \sum_{\mathbf{U}} \lambda_X \Pr(X|\mathbf{U}) \prod_{Z \neq Y} M_{ZX}$$

where $Z$ ranges over the neighbors of $X$ and $\alpha$ is a normalizing constant.[4] Similarly, the message $X$ sends to a parent $U$ is

$$M_{XU} = \alpha \sum_{X\mathbf{U}-\{U\}} \lambda_X \Pr(X|\mathbf{U}) \prod_{Z \neq U} M_{ZX}.$$

Message passing continues until the message values converge. The posterior of $X$ is then approximated as

$$\Pr'(X|e) = \alpha \sum_{\mathbf{U}} \lambda_X \Pr(X|\mathbf{U}) \prod_{Z} M_{ZX}.$$

---

[4] We use potential notation more common to join trees than the standard descriptions of belief propagation because we believe the many indices required in standard presentations mask the simplicity of the algorithm.

### 3.2 Description of the Algorithm

We use BP for the inference algorithm, and stochastic hill climbing as the optimization routine. The stochastic hill climbing method performs local search in the space of MAP variable instantiations, looking for the optimal instantiation. It works by either greedily moving to the best neighbor, or stochastically selecting a neighbor, where the choice is made randomly with some fixed probability. In this algorithm, one instantiation of the MAP variables is a neighbor of another if they differ only in the assignment of a single variable.

The score for a particular instantiation can be computed using the method for approximating the probability of evidence given in [21]. Using this method to select the best neighbor to move to requires running belief propagation separately on each neighbor in order to compute its score. We can do better than that by running belief propagation on the current state $\mathbf{s}$, and using the messages to approximate the change in score that moving to a neighboring state $x, \mathbf{s} - X$ would produce. The improvement from the current state $\mathbf{s}$ to the neighboring state $x, \mathbf{s} - X$ is just the ratio of their probabilities

$$\text{improvement}(x, \mathbf{s} - X) = \frac{\Pr'(x, \mathbf{s} - X, \mathbf{e})}{\Pr'(\mathbf{s}, \mathbf{e})}.$$

By dividing both numerator and denominator by $\Pr'(\mathbf{s} - X, \mathbf{e})$, we get

$$\text{improvement}(x, \mathbf{s} - X) = \frac{\Pr'(x|\mathbf{s} - X, \mathbf{e})}{\Pr'(x_\mathbf{s}|\mathbf{s} - X, \mathbf{e})}$$

where $x_\mathbf{s}$ is the value that $X$ takes on in $\mathbf{s}$. So, given the ability to approximate retracted conditional probabilities locally, we can compute the best neighbor after a single belief propagation.

Belief propagation is able to approximate retracted values for each variable efficiently based on the messages passed to that variable. For polytrees, the incoming messages are independent of the value of the local CPT or any evidence entered. Leaving the evidence out of the product yields

$$\Pr(X|\mathbf{e} - X) = \alpha \sum_{\mathbf{U}} \Pr(X|\mathbf{U}) \prod_{Z} M_{ZX}.$$

In multiply connected networks the incoming messages are not necessarily independent of the evidence or the local CPT, but as is done with other BP methods, we ignore that and hope that it is nearly independent. Empirically, the approximation seems to be quite accurate. Figure 4 shows a representative example, comparing the correspondence between the approximate and exact retracted probabilities for 30 variables in the Barley network. The x axis corresponds to the true retracted probability, and the y axis to the approximation produced using belief propagation.

Using retracted conditional probabilities to compute the improvement provides a linear speedup as compared to using belief propagation to compute the score for each neighbor separately. Figure 5 provides pseudocode for the algorithm.

### 3.3 Initializing the Search

The performance of local search methods such as hill climbing often depend crucially on the initialization. We investigate two methods previously shown to be successful when using exact inference. The first method is based on MPE. It consists of computing the MPE assignment (which we approximate using the standard BP approximation method) then creating the MAP assignment by setting each MAP variable to the value it takes on in the MPE assignment. The other method creates the instantiation by setting each MAP variable to the instance that maximizes $\Pr'(X|e)$, which we will call ML.

### 3.4 Experimental Results

We tested the algorithm on both synthetic and two real world networks from the Bayesian network repository [1]. For the first experiment, we generated 100 synthetic networks with 100 variables each using the method described in [15] with bias parameter 0.25 and width parameter of 13. We generated the networks to be small enough that we could often compute the exact MAP value, but large enough to make the problem challenging. We chose the MAP variables as the roots (typically between 20 and 25 variables), and the evidence values were chosen randomly from 10 of the leaves. We computed the true MAP for the ones which memory constraints (512 MB of RAM) allowed. We computed the true probability of the instantiations produced by the two initialization methods. We also computed the true probability of the instantiations returned by pure hill climbing (i.e. only greedy steps were taken), and stochastic hill climbing (using $p_f = .3$ and 100 iterations) for both initialization methods. Of the 100 networks, we were able to compute the exact MAP in 59 of them. Table 1 shows the number exactly solved for each method, as well as the worst instantiation produced, measured as the ratio of the probabilities of the found instantiation to the true MAP instantiation. All of the hill climbing methods improved significantly over their initializations in general, although for 2 of the networks, the hill climbing versions were slightly worse than the initial value (the worst was a ratio of .835), because of a slight mismatch in the true vs. approximate probabilities. Over-all, the stochastic hill climbing routines outperformed the other methods.

|  | # solved exactly | worst |
|---|---|---|
| MPE | 9 | .015 |
| MPE-Hill | 41 | .06 |
| MPE-SHill | 43 | .21 |
| ML | 31 | .34 |
| ML-Hill | 38 | .46 |
| ML-SHill | 42 | .72 |

Table 1: Solution quality for the random networks. Shows the number solved exactly of the 59 for which we could compute the true MAP value. Worst is the ratio of the probabilities of the found instantiation to the true MAP instantiation. Each hill climbing method improved significantly over the initializations.

|  | min | median | mean | max |
|---|---|---|---|---|
| MPE-Hill | 1.0 | 8.4 | $1.3 \times 10^{11}$ | $3.1 \times 10^{12}$ |
| MPE-SHill | 1.0 | 8.4 | $1.3 \times 10^{11}$ | $3.1 \times 10^{12}$ |
| ML-Hill | $1.0 \times 10^4$ | $3.6 \times 10^7$ | $3.4 \times 10^{15}$ | $8.4 \times 10^{16}$ |
| ML-SHill | $7.7 \times 10^3$ | $3.6 \times 10^7$ | $3.4 \times 10^{15}$ | $8.4 \times 10^{16}$ |

Table 2: The statistics on the improvement over just the initialization method for each search method on the data set generated from the Barley network. Improvement is measured as the ratio of the found probability to the probability of the initialization instantiation.

In the second experiment, we generated 25 random MAP problems for the Barley network, each with 25 randomly chosen MAP variables, and 10 randomly chosen evidence assignments. We use the same parameters as in the previous experiment. The problems were to hard to compute the exact MAP, so we report only on the relative improvements over the initialization methods. Table 2 summarizes the results. Again, the stochastic hill climbing methods were able to significantly improve the quality of the instantiations created.

In the third experiment, we performed the same type of experiment on the Pigs network. None of the search methods were able to improve on ML initialization. We concluded that the problem was too easy. Pigs has over 400 variables, and it seemed that the evidence didn't force enough dependence among the variables. We ran another experiment with Pigs, this time using 200 MAP variables and 20 evidence values to make it more difficult. Table 3 summarizes the results. Again, the stochastic methods were able to improve significantly over the initialization methods.

```
Given: Bayesian network N, evidence e, and MAP variables S.
Compute: An instantiation s which (approximately) maximizes Pr(s, e).

Initialize current state s to some instantiation of S.
s_best = s
Repeat many times:
    Perform belief propagation on N with evidence s, e.
    if Pr'(s, e) > Pr'(s_best, e) then
        s_best = s
    With probability p_f do
        Randomly modify one of the variable assignments in s.
    otherwise do
        Compute improvement(x, s − X) = Pr'(x|s − X, e)/ Pr'(x_s|s − X, e) for each neighbor x, s − X.
        if improvement(x, s − X) < 1 for all neighbors
            Randomly modify one of the variable assignments in s.
        else
            Set s to the neighbor x, s − X that has the highest improvement.
return s_best
```

Figure 5: An algorithm to approximate MAP using stochastic hill climbing and belief propagation

|          | min  | median             | mean               | max                |
|----------|------|--------------------|--------------------|--------------------|
| MPE-Hill | 1.0  | $1.7 \times 10^5$  | $1.5 \times 10^7$  | $3.3 \times 10^8$  |
| MPE-SHill| 1.0  | $2.5 \times 10^5$  | $4.5 \times 10^{11}$ | $1.1 \times 10^{13}$ |
| ML-Hill  | 13.0 | $2.0 \times 10^3$  | $3.3 \times 10^5$  | $4.5 \times 10^6$  |
| ML-SHill | 13.0 | $1.2 \times 10^4$  | $8.2 \times 10^5$  | $8.2 \times 10^6$  |

Table 3: The statistics on the improvement over just the initialization method alone for each search method on the data set generated from the Pigs network. Improvement is measured as the ratio of the found probability to the initialization probability.

## 4 Conclusion

MAP is a computationally very hard problem which is not in general amenable to exact solution even for very restricted classes (ex. polytrees). Even approximation is difficult. Still, we can produce approximations that are much better than those currently used by practitioners (MPE, ML) through using approximate optimization and inference methods. We showed one method based on belief propagation and stochastic hill climbing that produced significant improvements over those methods, extending the realm for which MAP can be approximated to networks that work well with belief propagation.

## Acknowledgement

This work has been partially supported by MURI grant N00014-00-1-0617

## References

[1] Bayesian network repository. www.cs.huji.ac.il/labs/compbio/Repository.

[2] A. Darwiche. Recursive conditioning. *Artificial Intelligence*, 126(1-2):5–41, February, 2001.

[3] L. de Campos, J. Gamez, and S. Moral. Partial abductive inference in Bayesian belief networks using a genetic algorithm. *Pattern Recognition Letters*, 20(11-13):1211–1217, 1999.

[4] R. Dechter. Bucket elimination: A unifying framework for probabilistic inference. In *12th Conference on Uncertainty in Artificial Intelligence*, pages 211–219, 1996.

[5] F. V. Jensen, S. L. Lauritzen, and K. G. Olesen. Bayesian updating in recursive graphical models by local computation. *Computational Statistics Quarterly*, 4:269–282, 1990.

[6] S. L. Lauritzen and D. J. Spiegelhalter. Local computations with probabilities on graphical structures and their application to expert systems. *Journal of Royal Statistics Society, Series B*, 50(2):157–224, 1988.

[7] M. L. Litmman, S. M. Majercik, and T. Pitassi. Stochastic boolean satisfiability. *Journal of Automated Reasoning*, 27(3):251–296, 2001.

[8] M. Littman. Personal comunication.

[9] M. Littman. Initial experiments in stochastic satisfiability. In *Sixteenth National Conference on Artificial Intelligence*, pages 667–672, 1999.

[10] M. Littman, J. Goldsmith, and M. Mundhenk. The computational complexity of probabilistic planning. *Journal of Artificial Intelligence Research*, 9:1–36, 1998.

[11] R. J. McEliece, E. Rodemich, and J. F. Cheng. The turbo decision algorithm. In *33rd Allerton Conference on Communications, Control and Computing*, pages 366–379, 1995.

[12] K. P. Murphy, Y. Weiss, and M. I. Jordan. Loopy belief propagation for approximate inference: an emperical study. In *Proceedings of Uncertainty in AI*, 1999.

[13] C. Papadimitriou. *Computational Complexity*. Addison-Wesley, Reading, MA, 1994.

[14] C. Papadimitriou and J. Tsitsiklis. The complexity of Markov decision processes. *Mathematics of Operations Research*, 12(3):441–450, 1987.

[15] J. D. Park and A. Darwiche. Approximating map using local search. In *17th Conference on Uncertainty in Artificial Intelligence,* pages 403–410, 2001.

[16] J. Pearl. *Probabalistic Reasoning In Intelligent Systems*. Morgan Kaufmann, 1998.

[17] D. Roth. On the hardness of approximate reasoning. *Artificial Intelligence*, 82(1-2):273–302, 1996.

[18] P. Shenoy and G. Shafer. Propagating belief functions with local computations. *IEEE Expert*, 1(3):43–52, 1986.

[19] S. E. Shimony. Finding maps for belief networks is NP-hard. *Artificial Intelligence*, 68(2):399–410, 1994.

[20] S. Toda. PP is as hard as the polynomial-time hierarchy. *SIAM Journal of Computing*, 20:865–877, 1991.

[21] Y. Weiss. Approximate inference using belief propagation. *UAI tutorial*, 2001.

[22] J. Yedidia, W. Freeman, and Y. Weiss. Generalized belief propagation. In *NIPS*, volume 13, 2000.